\documentclass{article}

\usepackage[preprint,nonatbib]{neurips_2024}

\usepackage[utf8]{inputenc} 
\usepackage[T1]{fontenc}    
\usepackage{hyperref}       
\usepackage{url}            
\usepackage{booktabs}       
\usepackage{amsfonts}       
\usepackage{nicefrac}       
\usepackage{microtype}      
\usepackage{mathtools} 
\usepackage{amsmath} 
\usepackage{multirow}
\usepackage{graphicx}
\usepackage{wrapfig}
\usepackage[toc,page]{appendix}
\usepackage[table,xcdraw]{xcolor}

\begin{document}

\title{Simple Feedfoward Neural Networks are Almost All You Need for Time Series Forecasting}

%

\author{%
  Fan-Keng~Sun \\
  Department of EECS, MIT \\
  \texttt{fankeng@mit.edu} \\
  \And
  Yu-Cheng~Wu \\
  Department of EECS, MIT \\
  \texttt{yuchengwu@mit.edu} \\
  \And
  Duane S.~Boning \\
  Department of EECS, MIT \\
  \texttt{boning@mtl.mit.edu}
}

\maketitle


\begin{abstract}
Time series data are everywhere---from finance to healthcare---and each domain brings its own unique complexities and structures.
While advanced models like Transformers and graph neural networks (GNNs) have gained popularity in time series forecasting, largely due to their success in tasks like language modeling, their added complexity is not always necessary.
In our work, we show that simple feedforward neural networks (SFNNs) can achieve performance on par with, or even exceeding, these state-of-the-art models, while being simpler, smaller, faster, and more robust.
Our analysis indicates that, in many cases, univariate SFNNs are sufficient, implying that modeling interactions between multiple series may offer only marginal benefits.
Even when inter-series relationships are strong, a basic multivariate SFNN still delivers competitive results.
We also examine some key design choices and offer guidelines on making informed decisions.
Additionally, we critique existing benchmarking practices and propose an improved evaluation protocol.
Although SFNNs may not be optimal for every situation (hence the ``almost'' in our title) they serve as a strong baseline that future time series forecasting methods should always be compared against.
\end{abstract}

\section{Introduction} \label{sec:intro}
Time series data are simply defined as information collected over time, without any additional constraints, making them ubiquitous and highly diverse.
For instance, meteorologists analyze weather station data to forecast tomorrow's temperature, freight companies monitor traffic flow to optimize routing, utility companies track electricity usage to improve distribution, and financial firms study past prices to allocate capital more effectively.
Although all these examples involve time series data, their structures vary widely, which is why models tailored with specific domain knowledge often yield the best results.
Traditionally, research on advanced time series forecasting---excluding linear models and some widely-used non-linear approaches---has been organized by target application.
For instance, in weather forecasting, effective models emerged only after computers became powerful enough to solve complex thermodynamics and hydrodynamics equations~\cite{weather_background1,weather_background2,weather_background3}.
Later, neural network models further improved accuracy by leveraging pattern recognition rather than relying solely on first-principle physics~\cite{pangu,graphcast,gencast}.
However, these neural network models are typically tailored specifically for weather data and are not directly applicable to other time series domains.
The same holds true for areas like traffic forecasting~\cite{traffic_background2,traffic_background1,traffic_background3}, electricity demand forecasting~\cite{demand_background1,demand_background2}, and economic forecasting~\cite{economic_background2,economic_background1}.

However, with the advent of deep learning and neural networks (NNs), researchers quickly recognized their potential and began applying them across diverse data modalities such as images, text, and speech. Time series forecasting also benefited, especially after the introduction of LSTNet~\cite{LSTNet}, which not only showcased the promise of NN-based models but also provided valuable benchmark datasets. Following this, a variety of NN-based models~\cite{AGCRN,DSANet,Conv-T,DA-RNN,TPA} have been proposed to enhance forecasting performance, primarily focusing on short-term predictions---up to 24 time steps ahead.

Inspired by the success of Transformers in natural language processing (NLP), researchers started exploring Transformer-based approaches for long-term forecasting, targeting horizons of up to 720 time steps. The Informer~\cite{haoyietal-informer-2021} was the first to formalize this long-term forecasting task, and its benchmarking protocol has since led to a surge of Transformer-based models. Starting with Autoformer~\cite{autoformer} and FEDformer~\cite{fedformer}, these models achieved promising improvements through increasingly complex modules and architectures.

Although one study~\cite{dlinear} questioned the effectiveness of Transformer-based models---suggesting that linear models might perform better---subsequent work, including PatchTST~\cite{patchtst}, PDF~\cite{pdf}, iTransformer~\cite{liu2023itransformer}, and DUET~\cite{duet}, has demonstrated that self-attention can still be effective for long-term forecasting. Consequently, considerable effort continues to be devoted to designing ever more sophisticated NN architectures, driven by the belief that capturing intricate temporal and spatial non-linearity is crucial for performance.

In this work, we reveal that such sophistication is unnecessary---at least on the current benchmark datasets.
While linear models cannot capture the inherent non-linearity in time series data, overly complex solutions often lead to overfitting given the dataset limitations.
We demonstrate that a simple feedforward neural network (SFNN) can match or even surpass state-of-the-art models like iTransformer~\cite{liu2023itransformer} and DUET~\cite{duet}.
Our approach avoids handcrafted modules designed to capture different scales, frequencies, or pattern clusters, making the model easily replicable within minutes.
In most cases, we adopt a channel-independent strategy where a univariate SFNN is shared across all series in a dataset where inter-series dependencies are weak.
For datasets with strong inter-series dependencies, adding a simple series-wise mapping (to learn patterns at the same time step) is sufficient to greatly improve performance.

In summary, our contributions are as follows:
\begin{itemize}
    \item Introduce the SFNN architecture and demonstrate that a simple univariate model can achieve state-of-the-art performance in long-term forecasting---even when compared with sophisticated Transformer-based models.
    \item Develop a multivariate version of SFNNs for scenarios with strong inter-series dependencies.
    \item Ablation study of three critical design choices and offer practical guidelines for decision-making.
    \item Critique current benchmarking practices and propose an improved evaluation protocol.
\end{itemize}

\section{Preliminaries}

\subsection{Definitions}
In time series forecasting, we have an input matrix $\mathbf{X} = \{\mathbf{X}_1, \dots, \mathbf{X}_t, \dots, \mathbf{X}_T\} \in \mathbb{R}^{T \times N}$ representing $N$ series (or channels) sampled at the same rate at the same time for $T$ time steps, where $\mathbf{X}_t \in \mathbb{R}^N$ is the $t$-th sample.
The goal is to forecast the values of $\{\mathbf{X}_{t+1}, \dots, \mathbf{X}_{t+H}\}$ given all histories $\{\mathbf{X}_1, \dots, \mathbf{X}_t\}$, where $H$ is the forecast horizon.
Thus, this is a sequence-to-sequence task where a single example is of the form (target, input) = $(\{\mathbf{X}_{t+1}, \dots, \mathbf{X}_{t+H}\}, \{\mathbf{X}_1, \dots, \mathbf{X}_t\})$.
The model designer can freely decide how to feed the histories into the model, e.g., what subset to use because it is often impractical to retain or use all histories for every prediction.
Since newer histories are more informative, a common approach is to select a suitable look-back length $L$ of the most recent histories as the input to the model.

\subsection{Problem Statement}
Mathematically speaking, given the model $f$, we aim to find the model parameters $\theta$ that minimize the mean squared error (MSE):
\begin{equation}
    \text{MSE} = \sum_t \sum_{h=1}^H \lVert \mathbf{X}_{t+h} - \hat{\mathbf{X}}_{t+h} \rVert^2_2
\end{equation}
on the test set, where $\{\hat{\mathbf{X}}_{t+H}, \dots, \hat{\mathbf{X}}_t\} = f(\mathbf{X}_{t-1}, \dots, \mathbf{X}_{t-L}; \theta)$ is the model forecast with look-back length $L$.

\subsection{Common Practices} \label{sec:common_practices}
As with any machine learning task, fair comparisons require splitting the data into training, validation, and test sets.
However, unlike text or image tasks, time series data cannot be randomly divided---the data must be divided in temporal order to avoid data leakage and ensure realistic evaluation.
For example, the first $70\%$ of the data might be used for training, followed by the next $10\%$ for validation and the last $20\%$ for testing.
Additionally, since time series datasets are often relatively small and subject to significant distribution drift~\cite{ci_cd}, properly comparing different methods becomes even more challenging.

Consider a forecasting model tasked with predicting influenza cases.
Suppose we have data up to December 2020 and intend to deploy the model starting in January 2021.
Since we cannot ``peek'' into 2021 during development, we must designate a validation set to estimate its future accuracy.
The most rigorous approach is to employ out-of-sample (OOS) K-fold cross-validation~\cite{CV1,CV3,CV2}, which is necessary given the autocorrelated errors common in time series data~\cite{AdjAuto}. 
In this framework, the training data is partitioned into \(K+1\) sequential subsets; for each fold \(k\), the \((k+1)\)th subset serves as the validation set while the preceding subsets are used for training.

However, current practices often deviate from these rigorous methods.
In many cases, practitioners “peek” at the test set to select the best hyperparameters, and use only the last portion of the training data as a single validation set to select the model during a single run.
Although ``peeking'' is impractical in real-world scenarios, it is often used in results reported in the literature (as shown in Table~\ref{tab:best} and described in Section~\ref{sec:exp_peeking}), where the test set is publicly available.
To explain the issue of a single validation set, we use the influenza forecasting model as an example.
If one were to use 2020 as the sole validation set to select a model for deployment in 2021, the resulting performance might not be ideal because 2020 is an atypical year that does not accurately reflect typical model performance.
This is why out-of-sample K-fold cross-validation is generally preferred due to its robustness on model selection for future unseen data.
While these practices do not entirely invalidate the benchmarking process, it is important for readers and practitioners to recognize these issues and interpret benchmarking results in the literature with caution.

\begin{figure}[b]
  \centering
  \includegraphics[angle=270, width=0.95\textwidth]{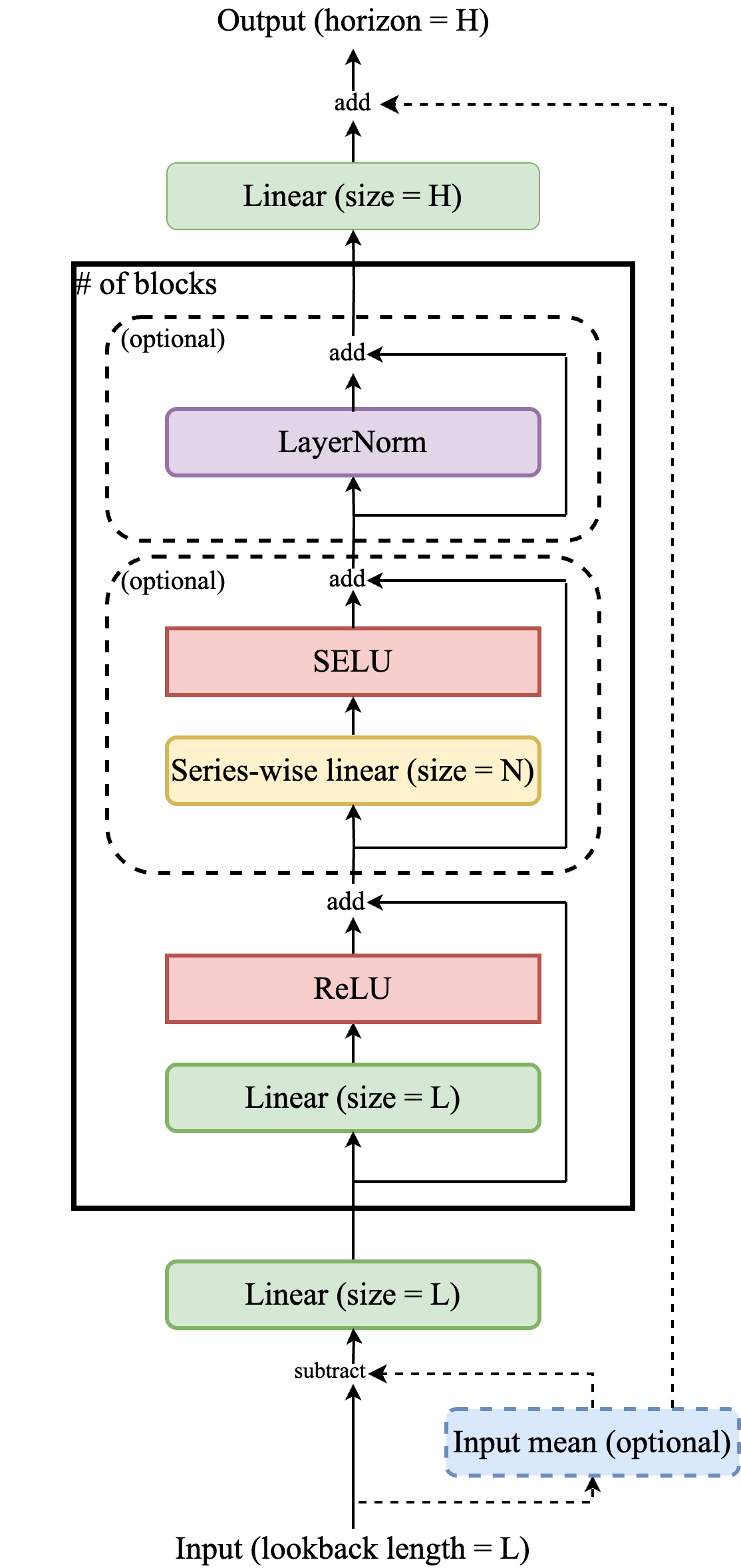}
  \caption{The SFNN architecture. The core block consists of linear mappings with ReLU activations and residual connections. Optional modules are dotted and include input mean centering, series-wise mapping with SELU activations, and layer normalization.}
  \label{fig:SFNN}
\end{figure}

\section{Simple Feedforward Neural Networks (SFNNs)}
As noted, our approach is exceptionally simple: Figure~\ref{fig:SFNN} captures the model details.
The key insight is that a univariate linear mapping shared across all series has previously proven relatively effective for forecasting~\cite{ci_cd,RevisitingLT,dlinear}.
To enhance its expressiveness, we add multiple inner blocks of linear plus ReLU (after the first linear layer) with residual connections that allows the model to incrementally adjust the output to capture more complex non-linear patterns.
This inner block may be repeated some number of times in a given SFNN.
Importantly, the model operates univariately and is shared across all series, meaning that for multivariate data, the same SFNN is applied to each series (each channel) during both training and inference.

This base version of SFNN is already quite powerful. However, three additional optional modules---input mean centering, series-wise non-linear mapping, and layer normalization---serve as the finishing touches, enabling SFNN to achieve state-of-the-art performance across a diverse range of time series datasets. These optional modules are described below.

\subsection{Input mean centering} \label{sec:centering}
First is the input mean centering module.
Originally, the input-output relationship is $\{\hat{\mathbf{X}}_{t+H}, \dots, \hat{\mathbf{X}}_t\} = f(\mathbf{X}_{t-1}, \dots, \mathbf{X}_{t-L}; \theta).$
We define $\hat{\mathbf{X}}_{\text{out}} \coloneqq \{\hat{\mathbf{X}}_{t+H}, \dots, \hat{\mathbf{X}}_t\}$ and $\mathbf{X}_{\text{in}} \coloneqq \{\mathbf{X}_{t-1}, \dots, \mathbf{X}_{t-L}\}$.
With input mean centering, the mean of the input $\bar{\mathbf{X}}_{\text{in}} \in \mathbb{R}^N$ is first subtracted from the input and then added back at the final output of the model: $\hat{\mathbf{X}}_{\text{out}} = f(\mathbf{X}_{\text{in}} - \bar{\mathbf{X}}_{\text{in}}; \theta) + \bar{\mathbf{X}}_{\text{in}}$.
When the data is multivariate, the input mean is calculated for each series separately.

This operation is designed to adapt to trends in the data.
While several previous studies have explored similar ideas, their approaches differ.
For example, works such as~\cite{duet,autoformer,dlinear,fedformer} decompose a trend component using a moving average kernel, and some, like~\cite{dlinear}, opt to use the last input value as a baseline.
We find that explicit trend decomposition is not necessary, and relying on the last input value actually diminishes performance in most benchmark cases.

\subsection{Series-wise non-linear mapping} \label{sec:mixer}
In most datasets, univariate SFNNs prove to be highly effective, indicating that explicitly modeling inter-series dependencies may offer little advantage.
However, in certain cases---most notably in the Solar Energy dataset---incorporating a series-wise non-linear mapping results in significant improvements.

\begin{table}[tbp]
\centering
\caption{Results by selecting the best performing lookback length by ``peeking'' at the testing set for comparing models. The mean-squared error (MSE) and the standard deviation of 10 runs are shown. Shaded numbers indicate the best performing models and is superscribed with $\dagger$ if the outperformance is statistically significant with $p$-value less than $5\%$.}
\label{tab:best}
{\scriptsize
\begin{tabular}{c|r||c|c|c}
Dataset           &  Horizon   & SFNN                & DUET~\cite{duet}                & iTransformer~\cite{liu2023itransformer}        \\
\hline
\hline
                               & 96  & $0.2790                        $ {\scriptsize $\pm 0.0003$} & $\cellcolor[HTML]{B7E1CD}0.2775^\dagger$ {\scriptsize $\pm 0.0008$} & $0.3027                        $ {\scriptsize $\pm 0.0014$} \\
                               & 192 & $\cellcolor[HTML]{B7E1CD}0.3161^\dagger$ {\scriptsize $\pm 0.0004$} & $0.3193                        $ {\scriptsize $\pm 0.0007$} & $0.3450                        $ {\scriptsize $\pm 0.0023$} \\
                               & 336 & $\cellcolor[HTML]{B7E1CD}0.3448$ {\scriptsize $\pm 0.0018$} & $0.3448                        $ {\scriptsize $\pm 0.0011$} & $0.3786                        $ {\scriptsize $\pm 0.0022$} \\
\multirow{-4}{*}{ETTm1}        & 720 & $\cellcolor[HTML]{B7E1CD}0.3814^\dagger$ {\scriptsize $\pm 0.0002$} & $0.3908                        $ {\scriptsize $\pm 0.0006$} & $0.4368                        $ {\scriptsize $\pm 0.0039$} \\
\hline
                               & 96  & $\cellcolor[HTML]{B7E1CD}0.1570^\dagger$ {\scriptsize $\pm 0.0007$} & $0.1627                        $ {\scriptsize $\pm 0.0013$} & $0.1781                        $ {\scriptsize $\pm 0.0021$} \\
                               & 192 & $\cellcolor[HTML]{B7E1CD}0.2113^\dagger$ {\scriptsize $\pm 0.0017$} & $0.2191                        $ {\scriptsize $\pm 0.0018$} & $0.2421                        $ {\scriptsize $\pm 0.0028$} \\
                               & 336 & $\cellcolor[HTML]{B7E1CD}0.2640$ {\scriptsize $\pm 0.0025$} & $0.2661                        $ {\scriptsize $\pm 0.0014$} & $0.2879                        $ {\scriptsize $\pm 0.0030$} \\
\multirow{-4}{*}{ETTm2}        & 720 & $0.3410                        $ {\scriptsize $\pm 0.0054$} & $\cellcolor[HTML]{B7E1CD}0.3378$ {\scriptsize $\pm 0.0031$} & $0.3753                        $ {\scriptsize $\pm 0.0090$} \\
\hline
                               & 96  & $\cellcolor[HTML]{B7E1CD}0.3503^\dagger$ {\scriptsize $\pm 0.0004$} & $0.3579                        $ {\scriptsize $\pm 0.0010$} & $0.3883                        $ {\scriptsize $\pm 0.0014$} \\
                               & 192 & $\cellcolor[HTML]{B7E1CD}0.3878^\dagger$ {\scriptsize $\pm 0.0019$} & $0.3964                        $ {\scriptsize $\pm 0.0043$} & $0.4216                        $ {\scriptsize $\pm 0.0019$} \\
                               & 336 & $\cellcolor[HTML]{B7E1CD}0.4119^\dagger$ {\scriptsize $\pm 0.0023$} & $0.4170                        $ {\scriptsize $\pm 0.0015$} & $0.4722                        $ {\scriptsize $\pm 0.0064$} \\
\multirow{-4}{*}{ETTh1}        & 720 & $\cellcolor[HTML]{B7E1CD}0.4355^\dagger$ {\scriptsize $\pm 0.0010$} & $0.4503                        $ {\scriptsize $\pm 0.0159$} & $0.5163                        $ {\scriptsize $\pm 0.0130$} \\
\hline
                               & 96  & $\cellcolor[HTML]{B7E1CD}0.2739$ {\scriptsize $\pm 0.0026$} & $0.2755                        $ {\scriptsize $\pm 0.0021$} & $0.3027                        $ {\scriptsize $\pm 0.0019$} \\
                               & 192 & $0.3368                        $ {\scriptsize $\pm 0.0034$} & $\cellcolor[HTML]{B7E1CD}0.3333^\dagger$ {\scriptsize $\pm 0.0012$} & $0.3761                        $ {\scriptsize $\pm 0.0026$} \\
                               & 336 & $0.3815                        $ {\scriptsize $\pm 0.0010$} & $\cellcolor[HTML]{B7E1CD}0.3641^\dagger$ {\scriptsize $\pm 0.0026$} & $0.4173                        $ {\scriptsize $\pm 0.0062$} \\
\multirow{-4}{*}{ETTh2}        & 720 & $0.3983                        $ {\scriptsize $\pm 0.0028$} & $\cellcolor[HTML]{B7E1CD}0.3831^\dagger$ {\scriptsize $\pm 0.0104$} & $0.4336                        $ {\scriptsize $\pm 0.0056$} \\
\hline
                               & 96  & $\cellcolor[HTML]{B7E1CD}0.1537^\dagger$ {\scriptsize $\pm 0.0025$} & $0.1647                        $ {\scriptsize $\pm 0.0013$} & $0.1844                        $ {\scriptsize $\pm 0.0041$} \\
                               & 192 & $\cellcolor[HTML]{B7E1CD}0.1832$ {\scriptsize $\pm 0.0018$} & $0.1864                        $ {\scriptsize $\pm 0.0044$} & $0.2000                        $ {\scriptsize $\pm 0.0025$} \\
                               & 336 & $\cellcolor[HTML]{B7E1CD}0.1925^\dagger$ {\scriptsize $\pm 0.0023$} & $0.1950                        $ {\scriptsize $\pm 0.0017$} & $0.2096                        $ {\scriptsize $\pm 0.0016$} \\
\multirow{-4}{*}{Solar Energy} & 720 & $\cellcolor[HTML]{B7E1CD}0.1959^\dagger$ {\scriptsize $\pm 0.0010$} & $0.2005                        $ {\scriptsize $\pm 0.0045$} & $0.2124                        $ {\scriptsize $\pm 0.0014$} \\
\hline
                               & 96  & $0.3484                        $ {\scriptsize $\pm 0.0002$} & $\cellcolor[HTML]{B7E1CD}0.3454$ {\scriptsize $\pm 0.0008$} & $0.3468                        $ {\scriptsize $\pm 0.0022$} \\
                               & 192 & $0.3697                        $ {\scriptsize $\pm 0.0025$} & $0.3682                        $ {\scriptsize $\pm 0.0025$} & $\cellcolor[HTML]{B7E1CD}0.3649^\dagger$ {\scriptsize $\pm 0.0028$} \\
                               & 336 & $0.3853                        $ {\scriptsize $\pm 0.0011$} & $0.3917                        $ {\scriptsize $\pm 0.0010$} & $\cellcolor[HTML]{B7E1CD}0.3835^\dagger$ {\scriptsize $\pm 0.0018$} \\
\multirow{-4}{*}{Traffic}      & 720 & $0.4310                        $ {\scriptsize $\pm 0.0008$} & $0.4366                        $ {\scriptsize $\pm 0.0038$} & $\cellcolor[HTML]{B7E1CD}0.4182^\dagger$ {\scriptsize $\pm 0.0031$} \\
\hline
                               & 96  & $\cellcolor[HTML]{B7E1CD}0.1259^\dagger$ {\scriptsize $\pm 0.0003$} & $0.1275                        $ {\scriptsize $\pm 0.0002$} & $0.1480                        $ {\scriptsize $\pm 0.0003$} \\
                               & 192 & $\cellcolor[HTML]{B7E1CD}0.1433^\dagger$ {\scriptsize $\pm 0.0003$} & $0.1467                        $ {\scriptsize $\pm 0.0002$} & $0.1644                        $ {\scriptsize $\pm 0.0009$} \\
                               & 336 & $\cellcolor[HTML]{B7E1CD}0.1568^\dagger$ {\scriptsize $\pm 0.0004$} & $0.1594                        $ {\scriptsize $\pm 0.0012$} & $0.1785                        $ {\scriptsize $\pm 0.0008$} \\
\multirow{-4}{*}{Electricity}  & 720 & $\cellcolor[HTML]{B7E1CD}0.1865                        $ {\scriptsize $\pm 0.0002$} & $0.1902$ {\scriptsize $\pm 0.0107$} & $0.2093                        $ {\scriptsize $\pm 0.0014$} \\
\hline
\hline
\multicolumn{2}{c||}{$1^\text{st}$ Count} & $19$ & $6$ & $3$ \\
\multicolumn{2}{c||}{$1^\text{st}$ Count with $p < 5\%$} & $14$ & $4$ & $3$  \\
\multicolumn{2}{c||}{Avg. loss rel. to SFNN} & $1.000$ & $1.011$ & $1.101$ 

\end{tabular}
}
\end{table}

Modeling inter-series dependencies is inherently more challenging than capturing temporal patterns.
While previous work on multivariate models has emphasized sophisticated methods to capture these relationships, our approach remains simple.
Our series-wise non-linear mapping employs the same residual connection concept but treats each time step as an independent univariate series, leading to a linear transformation represented by an \(N \times N\) matrix, followed by an activation.
Notably, we found that using the SELU activation function~\cite{selu} performs slightly better than ReLU in this setting.
Although the reason is not entirely clear, we suspect that SELU's ability to normalize different scales across series contributes to its improved performance.

\subsection{Layer normalization} \label{sec:layernorm}
Layer normalization~\cite{layernorm} has proven essential for stabilizing training in sequence modeling tasks, especially when combined with Transformer architectures~\cite{attention}.
As a result, many Transformer-based approaches, such as those in~\cite{liu2023itransformer,duet}, incorporate layer normalization.
We have observed that while layer normalization can be beneficial for some datasets, it may also lead to overfitting in others.
This observation is consistent with findings from~\cite{adanorm}, which demonstrated that the bias and gain parameters in layer normalization can contribute to overfitting.

\section{Experimental Results}
In this section, we present extensive experiments to validate the effectiveness of SFNNs.
First, we benchmark our method against previous state-of-the-art approaches, adhering to the common practices described in Section~\ref{sec:common_practices}.
Next, we conduct an ablation study of the three optional modules discussed in Sections~\ref{sec:centering}, \ref{sec:mixer}, and \ref{sec:layernorm}, and provide practical guidelines for selecting the appropriate modules based on application characteristics.
Finally, we critique current benchmarking practices from multiple perspectives and propose an improved evaluation protocol.

\renewcommand{\arraystretch}{1.2}
\begin{table}[tbp]
\centering
\caption{The look-back lengths $L$ that are scanned over for hyperparameter search for each dataset are multiples of the periods of the datasets. For Exchange rate, there is no clear periodicity so we use 5 because there are five trading days in a week.}
\label{tab:lookback_lengths}
\setlength{\tabcolsep}{2.5pt}
{\small
    \begin{tabular}{l|c|c|c|c|c|}
    & ETTm1/ & ETTh1/ETTh2/ & Solar/ & & \\
    \multirow{-2}{*}{Dataset(s)} & ETTm2 & Traffic/Electricity & Weather & \multirow{-2}{*}{ILI} & \multirow{-2}{*}{Exchange rate} \\
    \hline
    \hline
    Period & 96 & 24 or 168 & 144 & 52 & 5 \\
    \hline
    Look-back & 96, 192, 384, & 168, 336, & 144, 288, & 52, 104, & 5, 10, 20, 40, \\
    length $L$ & 672, 1344 & 672, 1344 & 576, 1008 & 208 & 80, 160, 320
    \end{tabular}
}
\end{table}
\renewcommand{\arraystretch}{1}

\subsection{Results following common practices} \label{sec:best_exp}
We compare SFNNs with several state-of-the-art methods using 10 widely adopted datasets from~\cite{LSTNet,autoformer,haoyietal-informer-2021} (see Appendix~\ref{app:datasets} for dataset statistics and descriptions).
Following previous work~\cite{haoyietal-informer-2021}, each dataset is first $z$-score normalized using the training set for each series separately and then fixed during training and testing.
In line with the common practices outlined in Section~\ref{sec:common_practices}, we ``peek'' at the testing set to select optimal hyperparameters. 
(We discuss results using the better cross-validation method in Section~\ref{sec:exp_peeking} and Table~\ref{tab:fair}.)
Given the large number of parameters to tune, we concentrate on the look-back length \(L\), a key hyperparameter shared across all models that significantly affects performance~\cite{moirai-moe,duet,moirai}.
This focus also underscores one of the critiques we discuss in Section~\ref{sec:crit_length}.
The range of look-back lengths we evaluated is provided in Table~\ref{tab:lookback_lengths}.

The results are presented in Table~\ref{tab:best}.
We include only seven datasets here, as the remaining three (ILI, Weather, and Exchange rate) have issues, as discussed in Section~\ref{sec:dataset_issues}.
We report only mean-squared error (MSE) rather than mean-absolute error (MAE) for several reasons: (1) MSE is arguably more practical in real-world applications since it penalizes larger errors more heavily; (2) using a different scoring function would require tuning separate hyperparameters, effectively doubling the computational effort; and (3) the results are largely similar, so we aim to keep the table concise.

Based on the results, SFNN outperforms both DUET~\cite{duet} and iTransformer~\cite{liu2023itransformer} on most datasets and forecasting horizons.
Specifically, SFNN was the best model in 19 out of 28 cases, with 14 of these improvements being statistically significant.
This indicates that, in many scenarios, SFNNs are almost all you need for time series forecasting.
However, limitations remain, particularly on the Traffic dataset.
We observed that while univariate SFNNs tend to slightly underfit the Traffic dataset, the multivariate variant overfits, leading to overall suboptimal performance.
A similar, though less pronounced, behavior is noted on the ETTh2 dataset.
These results suggest that the simple temporal or series-wise mapping employed by SFNNs may be insufficient for some datasets with distinct characteristics, which is not unexpected given the variability across different domains.

\begin{figure}[t]
  \centering
  \begin{minipage}{0.52\textwidth}
    \centering
    \includegraphics[width=\linewidth]{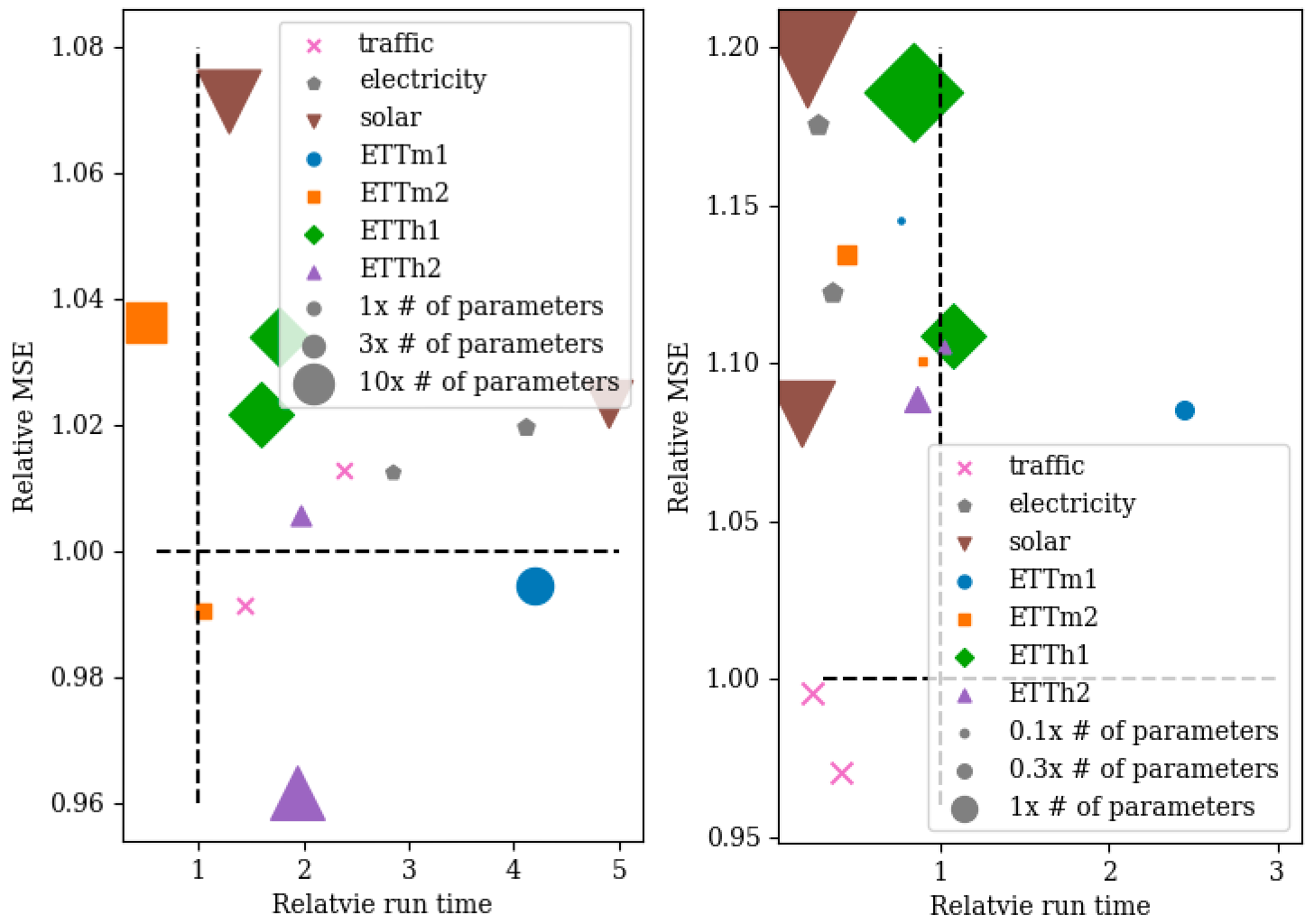}
    \caption{The relative run times, MSEs, and parameter counts of DUET~\cite{duet} and iTransformer~\cite{liu2023itransformer} compared to SFNNs on various datasets. For each dataset, horizon 96 and 720 are compared.}
    \label{fig:runtimes}
  \end{minipage}
  \hfill
  \begin{minipage}{0.47\textwidth}
    \centering
    \includegraphics[width=\linewidth]{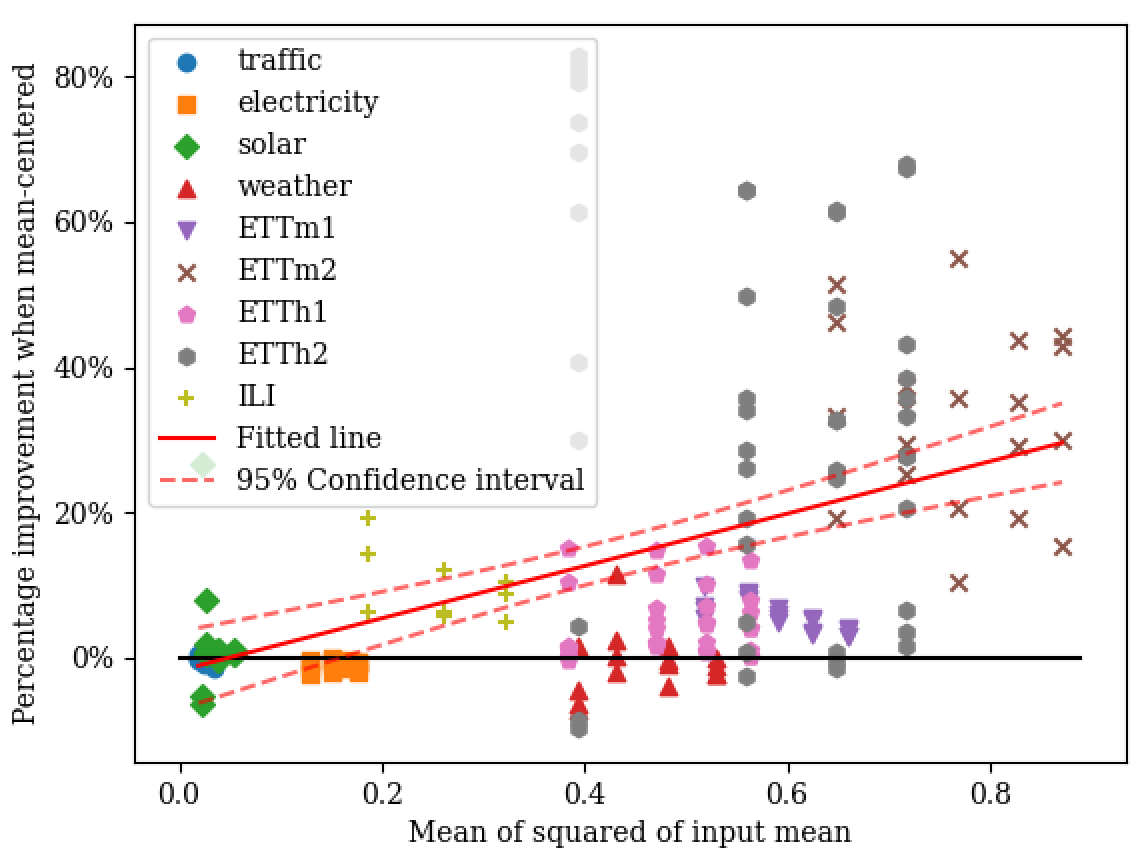}
    \vspace{-10pt}
    \caption{The percentage improvement by applying mean centering across all datasets and horizons, along with a linear regression line and its $95\%$ confidence interval.}
    \label{fig:centering}
  \end{minipage}
\end{figure}

\subsubsection{Model sizes and run times}
Figure~\ref{fig:runtimes} presents the relative run times, MSEs, and parameter counts of DUET~\cite{duet} and iTransformer~\cite{liu2023itransformer} compared to SFNNs on various datasets.
Because models converge at different rates, we define runtime as the total time taken to complete training.
In comparison, DUET~\cite{duet} typically features much larger model sizes and longer run times, although its performance is similar to that of SFNNs.
Conversely, iTransformer~\cite{liu2023itransformer} tends to have smaller model sizes and slightly less run times, but its performance is considerably worse.
Overall, these findings indicate that SFNNs strike an effective balance between model size, computational efficiency, and forecasting performance.

\subsection{Ablation study}
Our ablation study departs from the typical deep-learning approach of removing individual components to assess their contribution.
Given the extreme diversity of time series data, that method is not ideal.
Instead, we take a more statistical approach by correlating the improvements achieved by adding these components with a straightforward yet informative statistical property of the dataset.
This enables us to draw more robust, data-agnostic conclusions, supported by significance tests.

\subsubsection{Input mean centering}
Recall from Section~\ref{sec:centering} that input mean centering subtracts the mean from the input and then adds it back at the output, as described by the equation: $\hat{\mathbf{X}}_{\text{out}} = f(\mathbf{X}_{\text{in}} - \bar{\mathbf{X}}_{\text{in}}; \theta) + \bar{\mathbf{X}}_{\text{in}}.$
Despite its simplicity, this operation can lead to significant performance improvements.
For example, without mean centering, SFNNs would not achieve state-of-the-art performance on the ETT datasets.

\begin{figure}[t]
  \centering
  \includegraphics[width=\linewidth]{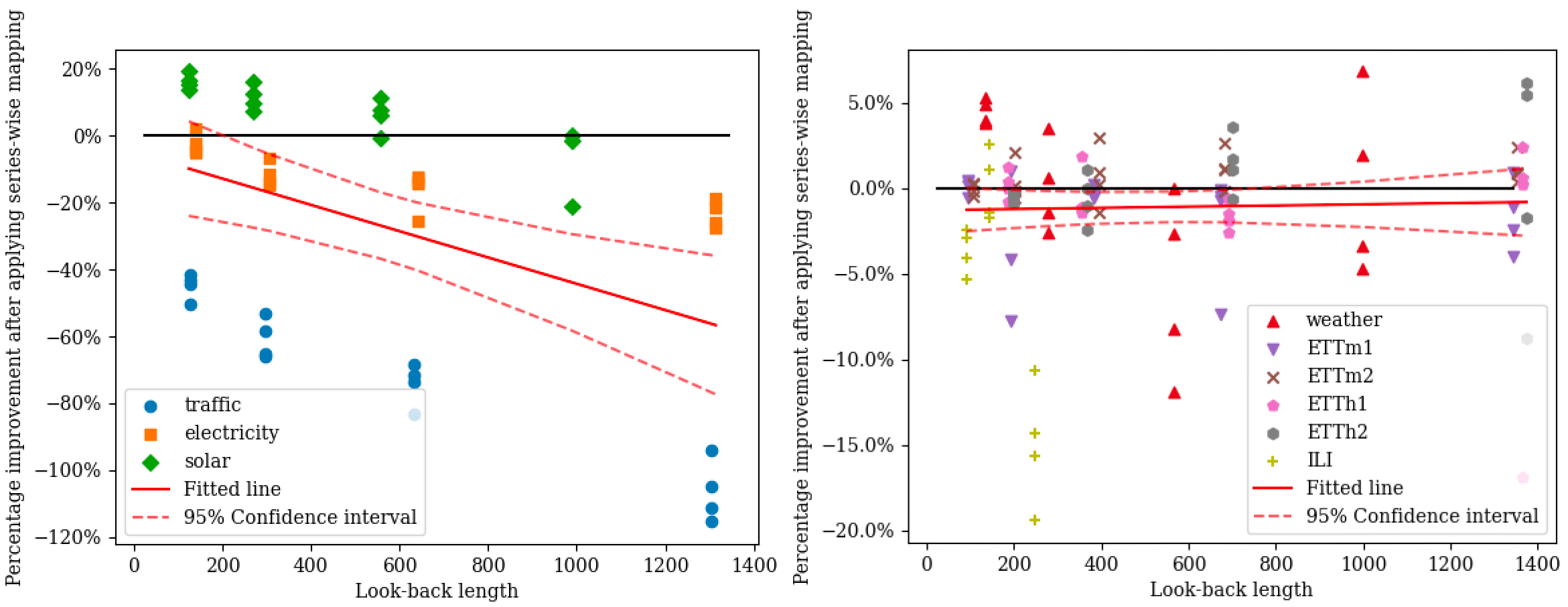}
  \caption{Percentage improvement from incorporating series-wise mapping, along with a linear regression line and its $95\%$ confidence interval. The left panel presents results for datasets with a large number of series, while the right panel shows those for datasets with a small number of series.}
  \label{fig:mixer}
\end{figure}

\begin{figure}[h]
  \centering
  \includegraphics[width=0.8\linewidth]{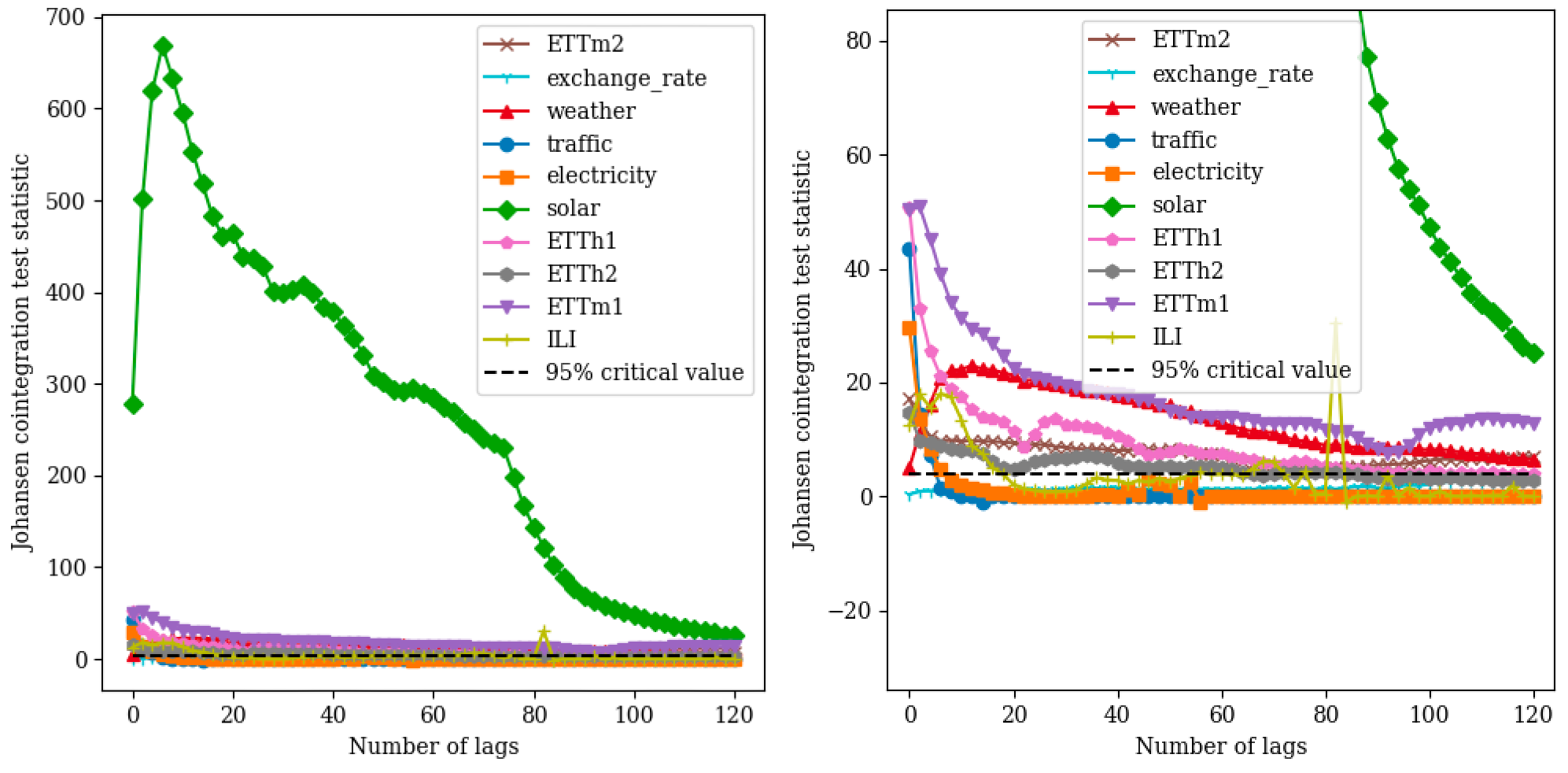}
  \caption{Johansen cointegration test results for all datasets, with the right panel offering a zoomed-in view. The horizontal axis represents the number of lags used in the test, while the vertical axis shows the test statistic for cointegration rank of \(r = N-1\). The dotted horizontal lines indicate the critical values for rejecting the null hypothesis with confidence of $95\%$ of cointegration rank of \(r = N-1\), which implies that all \(N\) series are cointegrated.}
  \label{fig:coint}
\end{figure}

Figure~\ref{fig:centering} displays the percentage improvement in terms of MSE for various combinations of dataset, forecast horizon, and look-back length. The \(x\)-axis represents the mean of the squared input mean across the entire training set:
\[
\text{mean of } \Big\{ \frac{1}{N} \|\bar{\mathbf{X}}_{t, \text{in}}\|_2^2, \forall t \in \text{training set} \Big\}.
\]
This value serves as a measure of the trend strength in the data because if a stronger trend exists, $\bar{\mathbf{X}}_{t, \text{in}}$ will deviate from zero more.

Figure~\ref{fig:centering} reveals a significantly positive correlation between the mean of the squared input mean and the improvement achieved through mean centering (although the relationship is slightly noisy due to dataset diversity). This indicates that when the trend in the data is stronger, applying mean centering is more beneficial. We also observe that when the mean of the squared input mean exceeds 0.2, mean centering is generally advantageous and is statistically significantly positive at $p$-value = $5\%$. However, given the diversity of the datasets, this should be regarded as a general guideline rather than a strict rule.

\subsubsection{Series-wise non-linear mapping}
Capturing inter-series dependencies is challenging, especially with powerful non-linear models like neural networks, which can easily overfit.
This risk is heightened by the fact that most time series datasets are relatively small---often with the number of data points comparable to the number of parameters.
In our approach, since the series-wise mapping employs a hidden size of $N$, the likelihood of overfitting increases significantly when the dataset contains a large number of series.

The results in Figure~\ref{fig:mixer} support our hypothesis.
As shown in the left panel, when the number of series is large (e.g., \(N > 100\)), adding series-wise mapping can lead to significant performance degradation---especially on the Traffic dataset, where the MSE can increase by over $100\%$.
Since Traffic has more series than Electricity, which in turn has more than Solar Energy, the degree of degradation follows this order.
Moreover, the overfitting issue becomes even more pronounced with longer look-back lengths.
In contrast, this phenomenon is generally not observed in datasets with a small number of series (e.g., \(N < 30\)), as illustrated in the right panel of Figure~\ref{fig:mixer}.

\begin{figure}[t]
  \centering
  \includegraphics[width=0.5\textwidth]{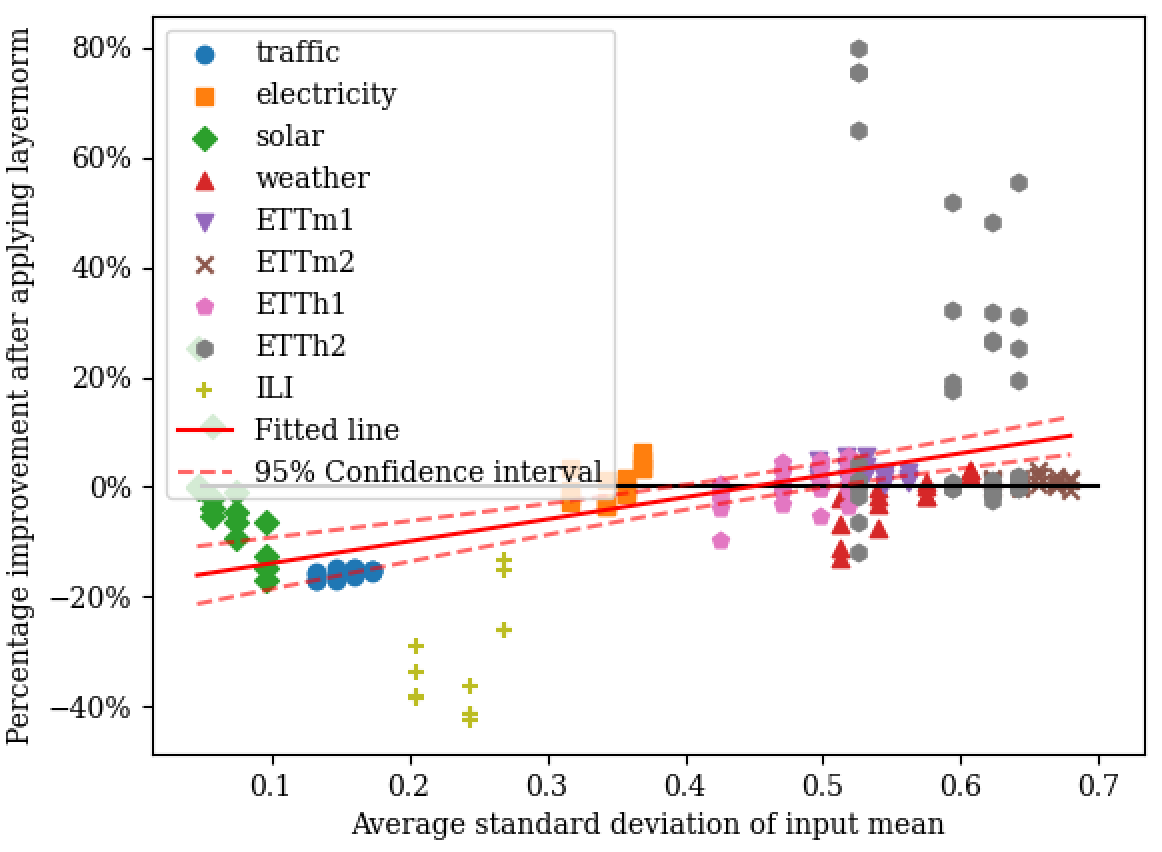}
  \caption{Percentage improvement after applying layer normalization on all datasets and horizons, along with a linear regression line and its $95\%$ confidence interval. The $x$-axis is the statistic indicating the scale different between series.}
  \label{fig:layernorm}
\end{figure}

However, one question remains: why is series-wise mapping so effective on the Solar Energy dataset where the performance can be improved by $20\%$ when the look-back length is around 200? Using the Johansen cointegration test~\cite{johansentest}, we uncover the reason.
This test assesses whether multiple time series share a long-term, stable relationship---known as cointegration.
As shown in Figure~\ref{fig:coint}, the Solar Energy dataset exhibits remarkably strong cointegration.
It is astonishing that 137 series remain statistically significantly cointegrated even with lags up to 120, but this makes sense given that the data measures photovoltaic production in one state (Alabama).
For most other datasets, significant cointegration persists over long lags so adding series-wise mapping is sometimes beneficial, except for the Traffic and Electricity datasets.
In these cases, cointegration falls below the critical value with only a short lag, indicating relatively weak long-term relationships among the series---an observation that aligns with the results in Figure~\ref{fig:mixer}.

\subsubsection{Layer normalization}
The final optional module of SFNNs is layer normalization, designed to stabilize training and reduce internal covariate shift by normalizing neuron activations~\cite{layernorm,adanorm}.
In each input matrix \(\mathbf{X}_{\text{in}} \in \mathbb{R}^{L \times N}\), the scales may vary over time or across series.
In our case, series-wise scale differences are more pronounced because: (1) each univariate series has already been \(z\)-score normalized; (2) temporal scale changes are generally limited within $L$ steps, whereas series-wise scale variations can be substantial; and (3) SFNN weights are shared across all series, so a series with a larger scale disproportionately affects the activations.
Based on these considerations, we quantify the series-wise scale difference by computing the average standard deviation of the input mean:
\begin{equation} \label{eq:layernorm}
    \text{Average} \big\{ \text{Std} \{ \bar{\mathbf{X}}_{t, \text{in}} \} \big\}, \forall t \in \text{training set},
\end{equation}
where $\bar{\mathbf{X}}_{t,\text{in}} \in \mathbb{R}^N$ denotes the mean of each series in the input.

Using Equation~(\ref{eq:layernorm}) as the \(x\)-axis, Figure~\ref{fig:layernorm} displays the impact of applying layer normalization.
As the \(x\)-axis increases---indicating a larger scale difference between series---the benefits of layer normalization also increase.
Additionally, within each dataset (with the exception of Solar, which exhibits strong cointegration), a similar trend is observed, albeit more weakly.
Based on the fitted line and the $95\%$ confidence interval, we can generally conclude that once the \(x\)-axis value exceeds $0.5$, incorporating layer normalization is usually beneficial. However, given the diversity of datasets, these guidelines should be applied with caution.

\subsection{Critiques of current benchmarking practices} \label{sec:critiques}

\subsubsection{Issues with the ILI, Weather, and Exchange rate datasets} \label{sec:dataset_issues}

\begin{table}[t]
\centering
\caption{Results for the ILI, Weather, and Exchange Rate datasets are based on selecting the overall best look-back lengths. The $N$ linears model fits \(N\) separate linear models, each with an individually tuned look-back length for each series. For the ILI and Weather datasets, these models also incorporate input mean centering. In the tables, the best-performing models are shaded, while the second-best are indicated in bold.}
\label{tab:problem_dataset}
{\scriptsize
\begin{tabular}{c|r||c|c|c|c}
Dataset           & Horizon    & $N$ linears & SFNN                & DUET~\cite{duet}                & iTransformer~\cite{liu2023itransformer}        \\
\hline
\hline

                      & 24 & $1.922$ & $\mathbf{1.6710                         \pm 0.0259}$ & $\cellcolor[HTML]{B7E1CD}1.6675 \pm 0.0615$ & $1.9931 \pm 0.1910$ \\
                      & 36 & $1.868$ & $\mathbf{1.6815                         \pm 0.0367}$ & $\cellcolor[HTML]{B7E1CD}1.6802 \pm 0.1122$ & $1.8731 \pm 0.1252$ \\
                      & 48 & $1.828$ & $\cellcolor[HTML]{B7E1CD}1.7336 \pm 0.0229$ & $\mathbf{1.7960                         \pm 0.1707}$ & $1.8491 \pm 0.1403$ \\
\multirow{-4}{*}{ILI} & 60 & $1.859$ & $\mathbf{1.8390                         \pm 0.0405}$ & $\cellcolor[HTML]{B7E1CD}1.7761 \pm 0.1440$ & $2.2390 \pm 0.1729$ \\
\hline
                          & 96  & $\cellcolor[HTML]{B7E1CD}0.1399$ & $\mathbf{0.1412 \pm 0.0007}$ & $0.1494                         \pm 0.0018$ & $0.1749 \pm 0.0014$ \\
                          & 192 & $\cellcolor[HTML]{B7E1CD}0.1821$ & $\mathbf{0.1874 \pm 0.0010}$ & $0.1887                         \pm 0.0010$ & $0.2248 \pm 0.0016$ \\
                          & 336 & $\cellcolor[HTML]{B7E1CD}0.2315$ & $0.2409                         \pm 0.0014$ & $\mathbf{0.2354 \pm 0.0009}$ & $0.2816 \pm 0.0011$ \\
\multirow{-4}{*}{Weather} & 720 & $\cellcolor[HTML]{B7E1CD}0.3025$ & $\mathbf{0.3040 \pm 0.0031}$ & $0.3075                         \pm 0.0022$ & $0.3584 \pm 0.0015$ \\
\hline
                          & 96  & $\cellcolor[HTML]{B7E1CD}0.0765$ & $\mathbf{0.0786 \pm 0.0007}$ & $0.0791 \pm 0.0008$ & $0.0815 \pm 0.0001$ \\
                          & 192 & $\cellcolor[HTML]{B7E1CD}0.1479$ & $\mathbf{0.1503 \pm 0.0003}$ & $0.1637 \pm 0.0033$ & $0.1693 \pm 0.0004$ \\
                          & 336 & $\cellcolor[HTML]{B7E1CD}0.2402$ & $\mathbf{0.2606 \pm 0.0023}$ & $0.2850 \pm 0.0086$ & $0.3093 \pm 0.0005$ \\
\multirow{-4}{*}{Exchange}& 720 & $\cellcolor[HTML]{B7E1CD}0.5487$ & $\mathbf{0.5770 \pm 0.0243}$ & $0.6593 \pm 0.0340$ & $0.8240 \pm 0.0016$
\end{tabular}
}
\end{table}

The ILI, Weather, and Exchange Rate datasets are commonly used for benchmarking, yet they provide limited insights into model performance.

The ILI dataset is particularly small, containing fewer than 1,000 time steps across seven series, and it shows significant distribution drift between the training and testing sets.
Although the values remain bounded during training and validation, several series exhibit a sharp increase during testing, with unforeseen patterns emerging in the final 20 steps due to the onset of COVID-19.
This is evident in Table~\ref{tab:problem_dataset}, where no model achieves an MSE lower than 1, a weak performance considering that the dataset is \(z\)-score normalized with respect to the training set.
Moreover, the extremely high standard deviation indicates that no model significantly outperforms the others, though it is worth noting that SFNNs exhibit much lower variability.
Therefore, caution is advised when interpreting results from the ILI dataset.

\begin{figure}[b]
  \centering
  \includegraphics[width=0.9\textwidth]{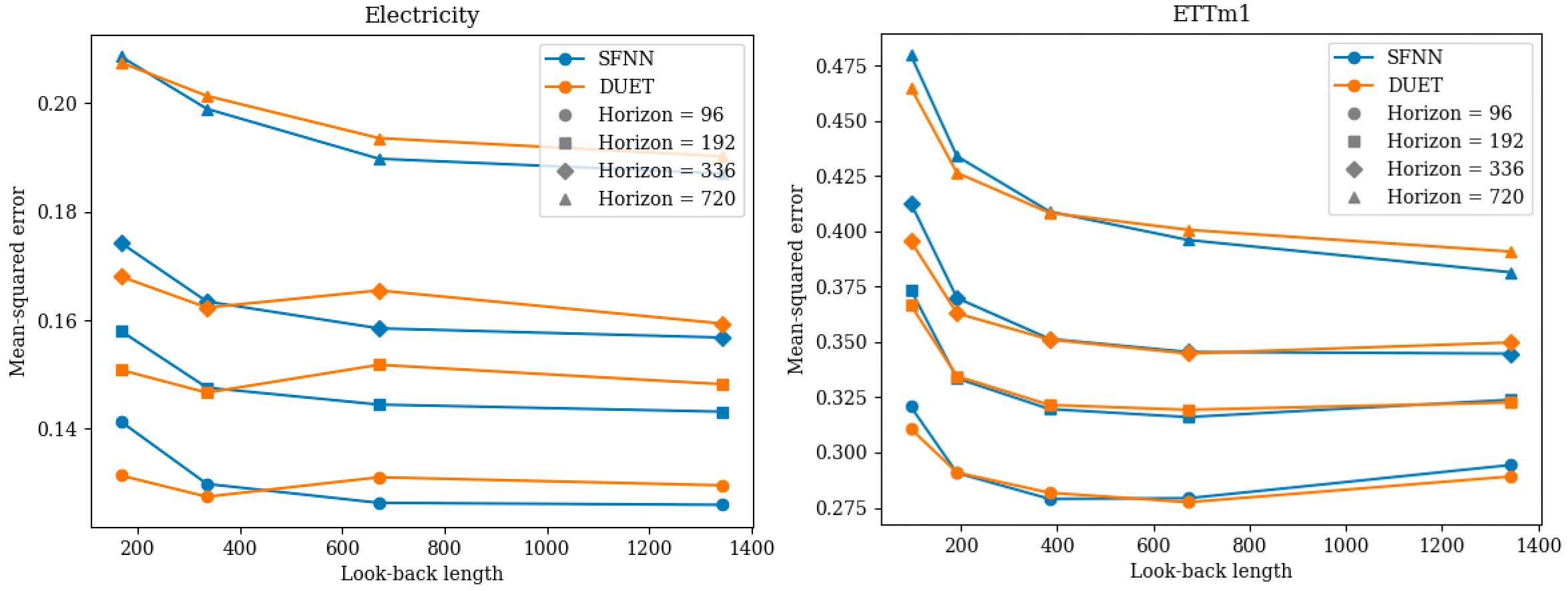}
  \caption{The results of SFNNs and DUETs on the Electricity (left) and ETTm1 (right) datasets with different horizons and look-back lengths.}
  \label{fig:lookback}
\end{figure}

For the Weather and Exchange Rate datasets, fitting \(N\) separate linear models---one per series---actually yields the best performance, as demonstrated in Table~\ref{tab:problem_dataset}.
This outcome suggests that these datasets behave more like \(N\) independent univariate series, making them unsuitable for benchmarking multivariate models, which tend to overfit.
In particular, the Weather dataset has been shown to be special in this regard, as confirmed by~\cite{ci_cd}.
Nonetheless, SFNNs still perform well on these datasets, ranking as the second-best model.

In summary, the ILI, Weather, and Exchange Rate datasets are too small and idiosyncratic to serve as reliable benchmarks for model performance.

\subsubsection{Affects of look-back length and randomness} \label{sec:crit_length}
In the early days of long-term forecasting, models were benchmarked with a fixed look-back length of 96, following the approach used in the original Informer paper~\cite{haoyietal-informer-2021}.
However, it has become clear that the look-back length is a critical hyperparameter that can dramatically affect performance.
Despite this, many recent studies~\cite{liu2023itransformer,duet,moirai,shi2025timemoe} still report results using a fixed look-back length of \(L = 96\).
This can be misleading because, in forecasting, all historical data is available, and different models may have different optimal look-back lengths.
Consider building a model to predict a sinusoidal wave with a period of \(P = 100\).
A linear model can achieve zero error when using a look-back length of \(L = P\) (since \(x_{t-100} = x_t\)), making it optimal.
However, if the look-back is restricted to only \(L = 3\), a non-linear model might outperform the linear one, even though it does not achieve the optimal performance.
Therefore, it is more reasonable to allow each model to determine its optimal look-back length.

\begin{table}[t]
\centering
\caption{Results by selecting the best performing look-back length using the validation set for comparing models. The mean-squared error (MSE) and the standard deviation of 10 runs are shown. Shaded numbers indicate the best performing models and is superscribed with $\dagger$ if the outperformance is statistically significant with $p$-value less than $5\%$.}
\label{tab:fair}
{\scriptsize
\begin{tabular}{c|r||c|c|c}
Dataset           & Horizon    & SFNN                & DUET~\cite{duet}                & iTransformer~\cite{liu2023itransformer}        \\
\hline
\hline
                               & 96  & $\cellcolor[HTML]{B7E1CD}0.2790^\dagger$  {\scriptsize $\pm 0.0003$} & $0.2909                        $ {\scriptsize $\pm 0.0008$} & $0.3027                        $ {\scriptsize $\pm 0.0014$} \\
                               & 192 & $\cellcolor[HTML]{B7E1CD}0.3196^\dagger$  {\scriptsize $\pm 0.0006$} & $0.3215                        $ {\scriptsize $\pm 0.0006$} & $0.3467                        $ {\scriptsize $\pm 0.0011$} \\
                               & 336 & $0.3514                                $  {\scriptsize $\pm 0.0001$} & $\cellcolor[HTML]{B7E1CD}0.3511$ {\scriptsize $\pm 0.0007$} & $0.3848                        $ {\scriptsize $\pm 0.0020$} \\
\multirow{-4}{*}{ETTm1}        & 720 & $\cellcolor[HTML]{B7E1CD}0.3960^\dagger$  {\scriptsize $\pm 0.0003$} & $0.4083                        $ {\scriptsize $\pm 0.0007$} & $0.4514                        $ {\scriptsize $\pm 0.0062$} \\
\hline
                               & 96  & $\cellcolor[HTML]{B7E1CD}0.1570^\dagger$  {\scriptsize $\pm 0.0007$} & $0.1657                        $ {\scriptsize $\pm 0.0017$} & $0.1781                        $ {\scriptsize $\pm 0.0021$} \\
                               & 192 & $\cellcolor[HTML]{B7E1CD}0.2139^\dagger$  {\scriptsize $\pm 0.0019$} & $0.2192                        $ {\scriptsize $\pm 0.0012$} & $0.2421                        $ {\scriptsize $\pm 0.0028$} \\
                               & 336 & $\cellcolor[HTML]{B7E1CD}0.2642^\dagger$  {\scriptsize $\pm 0.0012$} & $0.2661                        $ {\scriptsize $\pm 0.0014$} & $0.2879                        $ {\scriptsize $\pm 0.0030$} \\
\multirow{-4}{*}{ETTm2}        & 720 & $\cellcolor[HTML]{B7E1CD}0.3493^\dagger$  {\scriptsize $\pm 0.0015$} & $0.3563                        $ {\scriptsize $\pm 0.0032$} & $0.3761                        $ {\scriptsize $\pm 0.0042$} \\
\hline
                               & 96  & $\cellcolor[HTML]{B7E1CD}0.3503^\dagger$  {\scriptsize $\pm 0.0004$} & $0.3600                        $ {\scriptsize $\pm 0.0011$} & $0.4042                        $ {\scriptsize $\pm 0.0020$} \\
                               & 192 & $\cellcolor[HTML]{B7E1CD}0.3880^\dagger$  {\scriptsize $\pm 0.0005$} & $0.4091                        $ {\scriptsize $\pm 0.0005$} & $0.4489                        $ {\scriptsize $\pm 0.0087$} \\
                               & 336 & $\cellcolor[HTML]{B7E1CD}0.4138^\dagger$  {\scriptsize $\pm 0.0009$} & $0.4245                        $ {\scriptsize $\pm 0.0016$} & $0.4722                        $ {\scriptsize $\pm 0.0064$} \\
\multirow{-4}{*}{ETTh1}        & 720 & $\cellcolor[HTML]{B7E1CD}0.4355^\dagger$  {\scriptsize $\pm 0.0010$} & $0.5232                        $ {\scriptsize $\pm 0.0362$} & $0.7331                        $ {\scriptsize $\pm 0.0362$} \\
\hline
                               & 96  & $\cellcolor[HTML]{B7E1CD}0.2819^\dagger$  {\scriptsize $\pm 0.0005$} & $0.2846                        $ {\scriptsize $\pm 0.0022$} & $0.3027                        $ {\scriptsize $\pm 0.0019$} \\
                               & 192 & $\cellcolor[HTML]{B7E1CD}0.3588^\dagger$  {\scriptsize $\pm 0.0008$} & $0.3614                        $ {\scriptsize $\pm 0.0013$} & $0.3761                        $ {\scriptsize $\pm 0.0026$} \\
                               & 336 & $\cellcolor[HTML]{B7E1CD}0.3945        $  {\scriptsize $\pm 0.0020$} & $0.3962                        $ {\scriptsize $\pm 0.0040$} & $0.4173                        $ {\scriptsize $\pm 0.0062$} \\
\multirow{-4}{*}{ETTh2}        & 720 & $\cellcolor[HTML]{B7E1CD}0.4057^\dagger$  {\scriptsize $\pm 0.0027$} & $0.4171                        $ {\scriptsize $\pm 0.0030$} & $0.4336                        $ {\scriptsize $\pm 0.0056$} \\
\hline
                               & 96  & $0.1758                                $  {\scriptsize $\pm 0.0068$} & $\cellcolor[HTML]{B7E1CD}0.1647^\dagger$ {\scriptsize $\pm 0.0013$} & $0.1878                $ {\scriptsize $\pm 0.0052$} \\
                               & 192 & $\cellcolor[HTML]{B7E1CD}0.1846^\dagger$  {\scriptsize $\pm 0.0019$} & $0.1912                        $ {\scriptsize $\pm 0.0016$} & $0.2141                        $ {\scriptsize $\pm 0.0059$} \\
                               & 336 & $\cellcolor[HTML]{B7E1CD}0.1935^\dagger$  {\scriptsize $\pm 0.0023$} & $0.2078                        $ {\scriptsize $\pm 0.0013$} & $0.2263                        $ {\scriptsize $\pm 0.0028$} \\
\multirow{-4}{*}{Solar Energy} & 720 & $\cellcolor[HTML]{B7E1CD}0.1984^\dagger$  {\scriptsize $\pm 0.0019$} & $0.2077                        $ {\scriptsize $\pm 0.0015$} & $0.2124                        $ {\scriptsize $\pm 0.0014$} \\
\hline
                               & 96  & $\cellcolor[HTML]{B7E1CD}0.3484^\dagger$  {\scriptsize $\pm 0.0002$} & $0.3493                        $ {\scriptsize $\pm 0.0008$} & $0.3566                        $ {\scriptsize $\pm 0.0006$} \\
                               & 192 & $0.3697                                $  {\scriptsize $\pm 0.0025$} & $\cellcolor[HTML]{B7E1CD}0.3691$ {\scriptsize $\pm 0.0010$} & $0.3752                        $ {\scriptsize $\pm 0.0012$} \\
                               & 336 & $\cellcolor[HTML]{B7E1CD}0.3853^\dagger$  {\scriptsize $\pm 0.0011$} & $0.3922                        $ {\scriptsize $\pm 0.0003$} & $0.3891                        $ {\scriptsize $\pm 0.0011$} \\
\multirow{-4}{*}{Traffic}      & 720 & $0.4310                                $  {\scriptsize $\pm 0.0008$} & $0.4379                $ {\scriptsize $\pm 0.0027$} & $\cellcolor[HTML]{B7E1CD}0.4204^\dagger$ {\scriptsize $\pm 0.0042$} \\
\hline
                               & 96  & $\cellcolor[HTML]{B7E1CD}0.1261^\dagger$  {\scriptsize $\pm 0.0003$} & $0.1311                        $ {\scriptsize $\pm 0.0008$} & $0.1480                        $ {\scriptsize $\pm 0.0003$} \\
                               & 192 & $\cellcolor[HTML]{B7E1CD}0.1441^\dagger$  {\scriptsize $\pm 0.0002$} & $0.1467                        $ {\scriptsize $\pm 0.0002$} & $0.1645                        $ {\scriptsize $\pm 0.0010$} \\
                               & 336 & $\cellcolor[HTML]{B7E1CD}0.1586^\dagger$  {\scriptsize $\pm 0.0003$} & $0.1655                        $ {\scriptsize $\pm 0.0013$} & $0.1793                        $ {\scriptsize $\pm 0.0012$} \\
\multirow{-4}{*}{Electricity}  & 720 & $\cellcolor[HTML]{B7E1CD}0.1895^\dagger$  {\scriptsize $\pm 0.0007$} & $0.1936                        $ {\scriptsize $\pm 0.0010$} & $0.2112                        $ {\scriptsize $\pm 0.0052$} \\
\hline
\hline
\multicolumn{2}{c||}{$1^\text{st}$ Count} & 24 & 3 & 1 \\
\multicolumn{2}{c||}{$1^\text{st}$ Count with $p < 5\%$} & 23 & 1 & 1  \\
\multicolumn{2}{c||}{Avg. loss rel. to SFNN} & $1.000$ & $1.028$ & $1.117$ 
\end{tabular}
}
\end{table}

To illustrate, Figure~\ref{fig:lookback} plots the MSEs of SFNNs and DUETs on the Electricity and ETTm1 datasets across various horizons and look-back lengths.
The results show that although SFNNs may underperform with shorter look-back lengths, they outperform competitors when longer lags are used---indicating that SFNNs can more effectively leverage longer historical dependencies.

Besides the look-back length, randomness during training---stemming from factors such as initialization, batch shuffling, and dropout---can also impact performance, especially given the small size and distribution drift in many time series datasets.
Previous work typically reports a single result, which can mask variability and sometimes lead to statistically insignificant improvements.
In our experiments, we rigorously ran each experiment 10 times and report both the mean and standard deviation in Tables~\ref{tab:best}, \ref{tab:problem_dataset}, and \ref{tab:fair}, along with an assessment of statistical significance.

In summary, we advocate for a fairer benchmarking practice by enabling models to select their optimal look-back length and by reporting results across multiple trials to capture statistical significance.

\subsubsection{Reduce the effect of ``peeking''} \label{sec:exp_peeking}
As noted in Section~\ref{sec:common_practices}, a common issue with previous work~\cite{liu2023itransformer,duet,autoformer,haoyietal-informer-2021,fedformer} is ``peeking'' at the testing set to tune hyperparameters.
While it may be difficult to completely eliminate this practice, we believe there are more equitable approaches to mitigate its impact.
In theory, rigorous out-of-sample (OOS) K-fold cross-validation (as described in Section~\ref{sec:common_practices}) should be used to select the best hyperparameters.
However, this approach is often too complex and computationally intensive for widespread adoption.

Instead, we propose focusing on a single hyperparameter---the look-back length---based on the rationale outlined in Section~\ref{sec:best_exp}.
Rather than choosing the optimal look-back length by ``peeking'' at the testing set, as has been done in previous work and in our experiments in Section~\ref{sec:best_exp} and Table~\ref{tab:best}, we suggest selecting it using a validation set.
This approach better mimics real-world scenarios where the testing set is not available.
Moreover, if a model performs well under this procedure, it indicates greater robustness to hyperparameter variations and improved generalizability.
By incorporating the suggestions from Section~\ref{sec:crit_length}, the evaluation protocol becomes less biased.

The comparative results of SFNNs and other baseline models using this fairer procedure are summarized in Table~\ref{tab:fair}.
When comparing Tables~\ref{tab:best} and~\ref{tab:fair}, the performance advantage of SFNNs becomes even more pronounced, further supporting our claim that SFNNs are almost all you need for time series forecasting.

\section{Conclusion and Limitations}
In conclusion, our study demonstrates that simple feedforward neural networks (SFNNs) provide a surprisingly effective, robust, and efficient alternative for time series forecasting.
Despite their architectural simplicity, SFNNs consistently match or even exceed the performance of much more complex models.
Our statistical ablation studies further highlight the critical contributions of key components---such as input mean centering, series-wise mapping, and layer normalization---while offering practical guidelines for their implementation

Beyond evaluating model performance, we have critically examined existing benchmarking practices and proposed a fairer evaluation protocol.
By allowing models to select their optimal look-back length using the validation set and reporting results over multiple trials, our framework captures statistical significance and provides a more rigorous basis for comparing model performances.

Overall, our findings suggest that SFNNs are almost all you need for time series forecasting.
However, we do not claim that SFNNs are universally optimal; there are certainly datasets where specialized models may outperform them.
Nevertheless, SFNNs represent a strong baseline against which all future time series forecasting models should be compared.

\bibliography{references}
\bibliographystyle{abbrv}

\newpage
\appendix

\begin{appendices}
\section{Statistics and Descriptions of Datasets} \label{app:datasets}
\definecolor{Gray}{gray}{0.9}
\begin{table}[htb]
\centering
\setlength{\tabcolsep}{2.5pt}
{\scriptsize
    \begin{tabular}{l|cccc|cc}
    & Length & \# of series & Sampling & Data & Collection & \\
    \multirow{-2}{*}{Datasets} & $T$ & $N$ & period & split & year(s) & \multirow{-2}{*}{Description} \\
    \hline
    ETTm1 & 14,400 & 7 & 15 mins & 6:2:2 & 2016-17 & 6 power load features and oil temperatur of power transformer 1 \\
    \rowcolor{Gray}
    ETTm2 & 14,400 & 7 & 15 mins & 6:2:2 & 2016-17 & 6 power load features and oil temperatur of power transformer 2 collected \\
    ETTh1 & 57,600 & 7 & 1 hour & 6:2:2 & 2016-17 & Same as ETTm1 but down-sampled to hourly \\
    \rowcolor{Gray}
    ETTh2 & 57,600 & 7 & 1 hour & 6:2:2 & 2016-17 & Same as ETTm2 but down-sampled to hourly \\
    Solar & 52,560 & 137 & 10 mins & 7:1:2 & 2006 & Solar power production data from photovoltaic plants 
in Alabama State \\
    \rowcolor{Gray}
    Traffic & 17,544 & 862 & 1 hour & 7:1:2 & 2015-16 & Road occupancy rate (between 0 and 1) on Bay Area freeways \\
    Electricity & 26,304 & 321 & 1 hour & 7:1:2 & 2012-14 & Electricity consumption of 321 clients in kWh \\
    \rowcolor{Gray}
    ILI & 966 & 7 & 1 week & 7:1:2 & 2002-21 & Influenza-like illness (ILI) patients data from CDC \\
    Weather & 52,696 & 21 & 10 mins & 7:1:2 & 2020 & Meteorological indicators recorded from a weather station in Jena, Germany \\
    \rowcolor{Gray}
    Exchange & 7,588 & 8 & 1 day & 7:1:2 & 1990-16 & Exchange rate of currencies relative to the US Dollar
    \end{tabular}
}
\end{table}

\end{appendices}

\end{document}